\documentclass[letterpaper, 10 pt, conference]{ieeeconf}  

\IEEEoverridecommandlockouts                              

\usepackage{amsmath} 
\usepackage{cite}
\usepackage{amsfonts}
\usepackage{algorithmic}
\usepackage{graphicx}
\usepackage{textcomp}
\usepackage{caption}
\usepackage{subcaption} 
\usepackage{xcolor}
\usepackage{gensymb}
\usepackage{hyperref}

\usepackage[dvipsnames]{xcolor}
\usepackage[table]{xcolor}
\newcommand{\hlbox}[2]{%
  \begingroup
  \setlength{\fboxsep}{1pt}
  \colorbox{#1}{#2}%
  \endgroup
}

\definecolor{myred}{RGB}{255, 191, 191}
\definecolor{myorange}{RGB}{255, 223, 191}
\definecolor{myyellow}{RGB}{255, 250, 210}

\newcommand{\ours}{ROVER}

\title{\LARGE \bf
\ours{}: Regulator-Driven Robust Temporal Verification of Black-Box Robot Policies}

\author{Kristy Sakano, Jianyu An, Dinesh Manocha, Huan Xu%
\thanks{All authors are associated with the University of Maryland, College Park. Corresponding author: {\tt\small kvsakano@umd.edu}.
}}
\begin{document}


\maketitle
\thispagestyle{empty}   
\pagestyle{empty}

\begin{abstract}
We present a novel, regulator-driven approach for the temporal verification of black-box autonomous robot policies, inspired by real-world certification processes where regulators often evaluate observable behavior without access to model internals. Central to our method is a regulator-in-the-loop approach that evaluates execution traces from black-box policies against temporal safety requirements. These requirements, expressed as prioritized Signal Temporal Logic (STL) specifications, characterize behavior changes over time and encode domain knowledge into the verification process. We use Total Robustness Value (TRV) and Largest Robustness Value (LRV) to quantify average performance and worst-case adherence, and introduce Average Violation Robustness Value (AVRV) to measure average specification violation. Together, these metrics guide targeted retraining and iterative model improvement. Our approach accommodates diverse temporal safety requirements (e.g., lane-keeping, delayed acceleration, and turn smoothness), capturing persistence, sequencing, and response across two distinct domains (virtual racing game and mobile robot navigation). Across six STL specifications in both scenarios, regulator-guided retraining increased satisfaction rates by an average of 43.8\%, with consistent improvement in average performance (TRV) and reduced violation severity (LRV) in half of the specifications. Finally, real-world validation on a TurtleBot3 robot demonstrates a 27\% improvement in smooth-navigation satisfaction, yielding smoother paths and stronger compliance with STL-defined temporal safety requirements.

\end{abstract}


\section{INTRODUCTION}
Real-world certification processes for autonomous robots increasingly involve black-box systems whose internal models cannot be directly inspected by regulators.  Several challenges have been identified for certifying black-box autonomy, including difficulties in objectively measuring risk and the high redesign cost due to errors discovered late in the verification process \cite{clark2015tevv}. In safety-critical applications, failure of autonomous systems can cause severe harm to humans and property \cite{haRoboticAutonomousSystems2019, freundIntelligentAutonomousRobots1994, bogueGrowingUseRobots2018, thakkerBoldlyGoWhere2024}. Consequently, a key challenge is ensuring that black-box autonomous robots satisfy rigorous safety requirements; we refer to this process as {\em  black-box verification} \cite{Corso_2021}.


As correctness and safety of a robot's task execution emerge from its behavior over time, verification of black-box autonomous robots increasingly necessitates temporal verification techniques. These techniques reason about how behavior unfolds across sequences of states, such as persistence of safe behavior, timely responses, and correct ordering of required actions \cite{clarkeModelChecking2018, edviken_safety_nodate}. Existing methods typically aggregate statistics over trajectories (e.g., estimating failure rates) or examine single-state interactions (e.g., distance to obstacles) \cite{Moss_2024, Ji2025TandEofBlackBoxDrivingSystems}. These evaluations do not consider realistic temporal safety requirements that span time intervals and event sequences. For example, oscillatory behavior or a slow drift towards failure may be considered temporarily safe states, but do not adhere to temporal safety requirements. 



Signal Temporal Logic (STL) provides a natural formalism for verifying temporal safety requirements of autonomous robots\cite{roehmSTLModelChecking2016}. STL is a formal specification language for expressing temporal properties over continuous-time, real-valued signals, and is well-suited to dense-time systems such as mobile robots \cite{donzeSignalTemporalLogic2013}. Unlike aggregate metrics, STL specifications can distinguish brief, recoverable violations from sustained unsafe behavior. Most existing STL-based verification methods, however,  exploit internal structure access to perform model checking or reachability analysis over all possible trajectories, which limits their applicability to black-box controllers \cite{baierModelChecking}. To advance realistic certification processes, our goal is to apply STL to verify temporal safety requirements in black-box autonomous robots.


\textbf{Main Contributions:} In this work, we present \ours{} (\textbf{R}egulator-Driven r\textbf{O}bust \textbf{VER}ification of Black-Box Robot Policies), an approach for verifying black-box autonomous mobile robot policies against temporal safety specifications. Our approach treats the autonomy stack (perception, planning, and control software) as a completely opaque component, using only observed rollouts in the verification process. \ours{} provides targeted recommendations on where retraining or redesign is most needed to improve compliance with safety specifications. The novel contributions of our work include: 



\textbf{1. Advancing real-world certification processes:} 
We formalize human-readable temporal safety rules for lane keeping, acceleration, and turning behavior as STL specifications, enabling quantitative verification of rollout traces from the black-box policy while remaining independent of any internal controller access. Given these STL specifications, \ours{} computes robustness signals over each rollout and aggregates them into summary metrics that captures both satisfaction and violation severity over time. In contrast to existing black-box safety validation and simulation-based testing, which rely on surrogate safety measures or purely statistical failure estimates, \ours{} provides a formally grounded, trajectory-level assessment of learning-based black-box controllers.  



\textbf{2. Adaptability and robustness of \ours{} across distinct scenarios:} We evaluate \ours{} in two domains, a virtual racing game and an autonomous mobile robot platform. In each scenario, rollout traces from a reinforcement learning (RL) \textit{pre-verification model} are evaluated against STL specifications using Total Robustness Value (TRV), Largest Robustness Value (LRV), and Average Violation Robustness Value (AVRV). These robustness metrics guide targeted reward restructuring and retraining, producing a \textit{post-verification model} with consistently higher satisfaction rates and improved robustness metrics in both domains, demonstrating that \ours{} remains effectiveness under differing task dynamics and training setups.

\textbf{3. Targeted feedback for model improvement:} By incorporating domain expertise through prioritized specification weighting, \ours{} provides quantitative scores and qualitative feedback to guide policy retraining. In the absence of directly comparable STL-based verification methods for black-box robot policies, we evaluate \ours{} by measuring improvements over a pre-verification baseline: across six STL specifications in two scenarios, regulator-guided retraining increases specification satisfaction rates by an average of 43.8\%, improves TRV for all specifications, and achieves better (less negative) LRV and AVRV in three of six specifications. These results indicate that targeted retraining with \ours{} yields both higher average performance and reduced severity of temporal safety violations.

\section{Background and Related Works}

The breadth of verification techniques for autonomous systems encompasses field-testing, simulation-based testing, and methods derived from formal methods, among others \cite{leahy_grand_2024}. In this section, we mainly focus on {\em Formal Methods} (FM) verification techniques: mathematically rigorous approaches for the specification, development, and verification of cyber-physical systems \cite{clarkeModelChecking2018}. FM verification techniques allow designers to verify correctness and safety properties of controllers prior to real-world deployment. 

There is a wealth of work applying formal methods to robot policy learning for the synthesis of correct-by-construction robot policies \cite{aksaray2016qlearningrobustsatisfactionsignal, Jiang2021TemporalLogicRewardShapingRL, manganaris2026formalmethodsrobotpolicy, wongpiromsarnFormalSynthesisEmbedded2011a}. However, our focus is on the use of \textit{formal methods in robot policy verification}, i.e., verifying that an existing robot policy follows established behavior requirements, rather than enforcing restrictions during the synthesis of a new policy. Synthesis-oriented methods often assume access to and control over the training process, whereas our black-box setting involves fixed, often opaque policies available only through their input-output behavior. In this section, we review existing literature on formal verification of white-box and black-box autonomous robots.

\subsection{White-Box Verification}
A \textit{white-box} controller exposes an internal model, or has known controller dynamics that can be formally specified, enabling model-based analysis \cite{Corso_2021}. Existing techniques for evaluating white-box controllers include reachability analysis, model checking, and static property verification \cite{tranStarBasedReachabilityAnalysis2019, tranParallelizableReachabilityAnalysis2019}. However, these techniques typically assume full access to a discrete, tractable model and struggle with the scale and complexity of learning-based controllers. Discrete abstractions, required for tractable analysis, lose meaningful expressiveness in continuous, high-dimensional spaces, and are inadequate for capturing temporal sequence requirements of autonomous robots \cite{dimitrovaDeductiveControlSynthesis2014, lindemannControlBarrierFunctions2019}. Moreover, these FM techniques are not inherently designed to provide actionable feedback for controller re-training.

Another subclass of FM techniques for white-box verification is certificate-based methods, which provide mathematical guarantees by constructing functions (e.g., Lyapunov functions, barrier functions, or contraction metrics) that satisfy inequality bounds and ensure system stability within certain regions of the state space \cite{dawsonSafeControlLearned2022}. These methods offer strong mathematical guarantees but rely on analytic access to system dynamics or controller structure.

To evaluate controller behavior over time in white-box settings, \textit{Signal Temporal Logic} (STL) was introduced as a formalism for specifying temporal behaviors in continuous-time control systems \cite{donzeSignalTemporalLogic2013}. STL defines system specifications as combinations of atomic predicates (e.g., inequalities on signal values), Boolean operators, and bounded temporal operators such as globally ($G$), eventually ($F$), and until ($U$). Its quantitative semantics provide a robustness value that measures how strongly a trace satisfies or violates a specification, making STL attractive for verification and falsification of white-box control systems and learning-based controllers \cite{Donz2010RobustSO}. Recent applications of STL in white-box verification encode property constraints into satisfiability or optimization problems \cite{pulinaAbstractionrefinementApproachVerification2010} and use solvers to find violating counterexamples \cite{tjengEvaluatingRobustnessNeural2018, krishnamurthyDualApproachScalable2018}. Other applications include runtime assurance \cite{wongpiromsarnFormalMethodsAutonomous2023a} and STL-based shielding to filter unsafe actions \cite{WangMultiAgentSTLShielding}. However, these uses of STL are contingent on access to internal models or controller signals, which are not applicable in black-box controllers.

\subsection{Black-Box Verification}
A system is said to be a \textit{black-box} if its internal model is not known or is  too complex to reason about explicitly; only input-output behavior is observable \cite{Corso_2021}. In contrast to white-box verification methods, black-box verification must rely on observed executions rather than analytic models.

Current evaluation of deployed, learning-based black-box autonomous robots often relies on surrogate strategies that approximate or over-simplify real safety issues\cite{tafidisApplicationSurrogateSafety2023, wangReviewSurrogateSafety2021}. These include approaches that pre-define requirements through hand-crafted finite state machines \cite{costelloGeneratingCertificationEvidence2024}, run-time assurance mechanisms that supervise behavior with fallback safe actions \cite{costelloruntime2023}, and scenario-based testing where black-box controllers are subjected to parameterized environments to observe performance \cite{Thompson2023ASF}. Other approaches quantify risk statistically by estimating failure probabilities through repeated rollouts in simulation \cite{nordenEfficientBlackboxAssessment2020}. These approaches are often grouped under black-box safety validation or simulation-based testing for autonomous systems. Some recent work has begun exploring data-driven temporal verification for black-box controllers; for example, \cite{salamati2020datadrivenverificationsignaltemporal} develops a Bayesian verification framework for low-dimensional parameter systems with parametric uncertainty. However, their approach assumes restricted gray-box visibility of internal control functions, limiting applicability to high-dimensional, non-linear black-box controllers. 

These limitations highlight a verification gap: existing black-box verification techniques rarely provide formally grounded, trajectory-level evaluations of how behavior evolves over time. Our work addresses this temporal, trajectory-level verification gap by applying STL specifications and associated robustness measures (Section \ref{subsec:robustness}) to verify learning-based black-box autonomous mobile robot policies, complementing existing design-time policy synthesis and black-box simulation-based testing.





\section{Our Approach: \ours{}} \label{sec:approach}



\ours{} uses an iterative approach for verifying the performance of a learning-based, black-box robot policy against temporal safety requirements. A \textbf{Regulator} specifies temporal safety requirements and evaluates a black-box system, then passes design recommendations to the \textbf{Designer}, who updates the policy to better satisfy the specification. We walk through a single iteration of the evaluation process here.

\subsection{Modeling The System}
We begin by modeling the robot as a discrete-time dynamical system

\begin{equation}
x_{t+1} = f(x_t, u_t),
\end{equation}

\noindent where $x_t \in \mathcal{X}$ is the state and $u_t \in \mathcal{U}$ is the control input \cite{dawsonSafeControlLearned2022}. A learning-based controller, known as a robot policy, $\pi_\theta : \mathcal{X} \rightarrow \mathcal{U}$ with parameters $\theta$ induces the closed-loop behavior


\begin{equation}
x_{t+1} = f(x_t, \pi_\theta (x_t)).
\end{equation}

\noindent For a given initial condition $x_0$, the resulting trajectory (rollout trace) is

\begin{equation}
\tau = (x_0, x_1,\dots,x_T).
\end{equation}

\noindent We denote by $\mathcal{T}_\pi$ the (random) collection of traces generated by policy $\pi_\theta$.

\subsection{Regulator Specifications}\label{subsec:regulators}

We model the Regulator as an external governing authority that evaluates safety compliance without access to the internal policy $\pi_\theta$. The Regulator formalizes domain knowledge into a finite set of Signal Temporal Logic (STL) specifications

\begin{equation}
\Phi = \{\phi_1, \phi_2, \dots, \phi_m\},
\end{equation}

\noindent where each $\phi_i$ formalizes a regulator-defined requirement (e.g., safety, social compliance, mission constraint). For each $\phi_i$, a robustness function is defined over a trace $\tau$

\begin{equation}
\rho : \Phi \times \mathcal{T}_\pi \to \mathbb{R},
\end{equation}
\noindent where $\rho(\phi_i, \tau)$ denotes the STL robustness of trace $\tau$ with respect to $\phi_i$. Here $\rho(\phi_i, \tau) > 0$ indicates robust satisfaction, $\rho(\phi_i, \tau) = 0$ indicates boundary (exact) satisfaction, $\rho(\phi_i, \tau) < 0$ indicates a specification violation, and $|\rho(\phi_i, \tau)|$ measures the margin of satisfaction or violation \cite{Donz2010RobustSO}.

\subsection{Robustness Evaluation}\label{subsec:robustness}
In this section, we present our approach to evaluate robustness in a way that reflects task dynamics and training specifics. Given a policy $\pi_\theta$ and $m$ specifications, the Regulator draws $N$ rollout traces $\{\tau_k\}_{k=1}^N$ and, for each trace–specification pair, compute a robustness value $\rho(\phi_i,\tau_k)$, yielding an $N \times m$ matrix of robustness scores. For a fixed specification $\phi_i$, let

\begin{equation}
\mathcal{P}(\mathcal{T}_\pi, \phi_i) = {\rho_i^{(1)}, \rho_i^{(2)},\dots,\rho_i^{(N)}},
\end{equation}
\noindent denote the multiset of robustness values $\rho(\phi_i, \tau_k)$. We leverage existing STL metrics Total Robustness Value (TRV) and Largest Robustness Value (LRV) from  \cite{nesteriniMiningSpecificationsPredictive2025},

\begin{equation}
\begin{split}
TRV(\mathcal{T}_\pi, \phi_i) = \sum_{\rho \in \mathcal{P}(\mathcal{T}_\pi, \phi_i)}{\rho}, \\
LRV(\mathcal{T}_\pi, \phi_i) = \min_{\rho \in \mathcal{P}(\mathcal{T}_\pi, \phi_i)} \rho.
\end{split}
\end{equation}

\noindent To aggregate only violating traces, we extract

\begin{equation}
\mathcal{P}^-(\mathcal{T}_\pi, \phi_i) = \{\rho \in \mathcal{P}(\mathcal{T}_\pi, \phi_i) | \rho < 0\},
\end{equation}

\noindent and define Average Violation Robustness Voting (AVRV) as

\begin{equation}
AVRV (\mathcal{T}_\pi, \phi_i) = \frac{1}{\text{count}(\mathcal{P}^-(\mathcal{T}_\pi, \phi_i))} \sum_{\rho \in \mathcal{P}^- (\mathcal{T}_\pi, \phi_i)} \rho.
\end{equation}

\noindent Each metric thus defines a Regulator feedback map $\mathcal{M}_\phi$ as

\begin{equation}
\mathcal{M}_{\phi_i} = (TRV(\mathcal{T}_\pi, \phi_i), LRV(\mathcal{T}_\pi, \phi_i), AVRV(\mathcal{T}_\pi,\phi_i)).
\end{equation}

The TRV metric reflects the aggregated safety margin across all rollout traces, capturing the average robustness of the model's traces. A higher TRV indicates that most scenarios maintain a comfortable safety margin, suggesting reliability under normal operating conditions. In contrast, we interpret LRV as the single most critical instance of rule violation among the rollout traces, while AVRV measures the average severity of violations across failing traces.

Because robustness is not calibrated across different formulas, we interpret robustness on a per-rule basis: large negative values correspond to severe violations and values near zero indicate borderline satisfaction/violation. While positive values reflect comfortable satisfaction margins, our analysis focuses on negative robustness, which quantifies violation severity. We apply this interpretation consistently when analyzing these metrics across each scenario.

\subsubsection{Domain Experts}\label{subsec:domain_experts}

In many settings, not all specifications are equally critical: for example, collision avoidance is typically more important than rules pertaining to wear \& tear on the robot. To capture this,  Regulators or other domain experts assign an importance weight $w_i > 0$ to each specification $\phi_i$. These weights encode the relative criticality of the requirements. Using the weights and the previously defined Regulator feedback map $\mathcal{M}_{\phi_i} $, we can define a safety score $S(\pi)$ for policy $\pi_\theta$ as

\begin{equation}
  S(\pi) = \sum_{i=1}^{3}  w_i*\mathcal{M}_{\phi_i},
\end{equation}
which Designers can use to compare candidate policies and guide reward shaping or retraining. 

\subsubsection{Recommendations}\label{subsec:recommendations}
Beyond providing quantitative safety scores $S(\pi)$, the regulators can classify the safety profile of a policy using simple decision rules on the robustness metrics. For a given specification $\phi_i$, examples include:

\textbf{1. No recommendation (normal behavior):} if $TRV(\mathcal{T}_\pi, \phi_i)$, $LRV(\mathcal{T}_\pi, \phi_i)$, and $AVRV(\mathcal{T}_\pi, \phi_i)$ are near zero, the policy satisfies $\phi_i$ with small dispersion in robustness, and no action is required.
  
\textbf{2. Recommend policy improvement (systematic violations):} if $TRV(\mathcal{T}_\pi, \phi_i)$, $LRV(\mathcal{T}_\pi, \phi_i)$, and $AVRV(\mathcal{T}_\pi, \phi_i)$ are negative with a large magnitude, indicating frequent and/or severe violations, Regulators recommend policy improvement for the behaviors captured by $\phi_i$.
  
\textbf{3. Recommend edge-case analysis (rare but severe failures):} if $LRV(\mathcal{T}_\pi, \phi_i) << AVRV(\mathcal{T}_\pi, \phi_i)$, Regulators flag rare but catastrophic failures and recommend targeted analysis of edge case scenarios.

These recommendation categories are communicated to Designers alongside the safety score $S(\pi)$ for policy $\pi_\theta$, providing both quantitative scores and qualitative guidance from domain experts for subsequent model updates.

\subsection{Designer Formulation}

Designers have access to the policy $\pi_\theta$ and seek to optimize task performance subject to Regulator feedback. In our formulation, Designers treat $S(\pi)$ as a safety score and aim to select parameters $\theta$ that improve both task reward and safety adherence to the Regulator's temporal safety requirements. Using the regulator-provided scores and recommendations (Section~\ref{subsec:recommendations}), Designers decide whether to keep, retrain, or replace a given black-box policy.

Although these STL specifications can also guide robot policy synthesis, in many deployed settings Regulators do not control the underlying model architecture and must evaluate opaque, probabilistic policies whose responses can be difficult to predict. Our approach is therefore positioned on the regulatory side: it does not address coverage or rare-event simulation, nor does it synthesize new robot policies, but instead equips regulatory officials with temporal, robustness-based tools to verify the safety of black-box controllers. While our focus is on enabling Regulators to evaluate deployed black-box controllers, we believe the same STL-based specifications should also be made available to Designers to inform policy synthesis during training.



\subsection{Verification using STL}
In our application of \ours{}, the Designer trained a reinforcement learning (RL) policy for an autonomous mobile robot using Proximal Policy Optimization (PPO), using existing baseline implementations and treating resulting policy $\pi_\theta$ as a black-box controller \cite{Schulman2017ProximalPO, stable-baselines3, huang2022cleanrl}. \ours{} is agnostic to the specific black-box learning algorithm; in practice, a wide range of learning-based methods could be implemented, with the best choice determined during robot policy synthesis.

Human-defined safety rules are initially articulated in natural language and subsequently translated into Signal Temporal Logic (STL) specifications. In our approach this process is performed , though recent work has explored using large language models to convert natural language rules into temporal logic specifications \cite{coslerNl2specInteractivelyTranslating2023, englishVerifiableNaturalLanguage2025}. For each policy $\pi_\theta$, the Regulator collects $N$ rollout traces$\{\tau_k\}_{k=1}^N$ and evaluates robustness values $\rho(\phi_i,\tau_k)$ using TeLEx~\cite{Jha2017TeLExPS}, which are then aggregated into the TRV, LRV, and AVRV metrics and further combined into safety scores and recommendations as described in Section~\ref{sec:approach}. In this work, we focus on evaluating the temporal safety of the behaviors that are exercised by a given sampling procedure, and we do not address the orthogonal problem of designing coverage-optimal or rare-event-targeted sampling schemes.

To demonstrate the generality of this workflow, we apply \ours{} to two case studies: (1) a virtual racing game in the Mario Kart (SNES) environment (Section~\ref{sec:mariokart}), and (2) an autonomous mobile robot navigation scenario (Section~\ref{sec:turtlebots}). In each case study, a \textbf{pre-verification model} is evaluated using \ours{}, the reward structure is updated based on the resulting compliance metrics, and a \textbf{post-verification model} is re-evaluated to showcase improved adherence to the safety specifications. 


\section{Virtual Racing Game}\label{sec:mariokart}
\setcounter{secnumdepth}{2}

The virtual racing game scenario uses \textbf{Mario Kart (SNES)} \cite{stable-retro}, a stylized simulation engine that provides a controlled and reproducible platform for evaluating autonomous decision-making under safety specifications. In the notation of Section~\ref{sec:approach}, the trained driving policy corresponds to $\pi_\theta$ and the safety rules below define a subset of the STL specifications $\Phi$.

The model is trained on a full loop of the Mario Circuit 1 track. The reward function of the pre-verification model balances four objectives: checkpoint progression ($R_{cp}$), penalizing stationary behavior ($R_{motion}$), staying on drivable terrain ($R_{road}$), and minimizing lap time ($R_{time}$), illustrated in Eq. \ref{eq:mario_reward},

\begin{equation} \label{eq:mario_reward}
R_{pre} = R_{cp} + R_{motion} + R_{road} + R_{time}.
\end{equation}

\begin{figure}[h!]
  \centering
  \begin{subfigure}{0.46\linewidth}
    \centering
    \includegraphics[width=\linewidth]{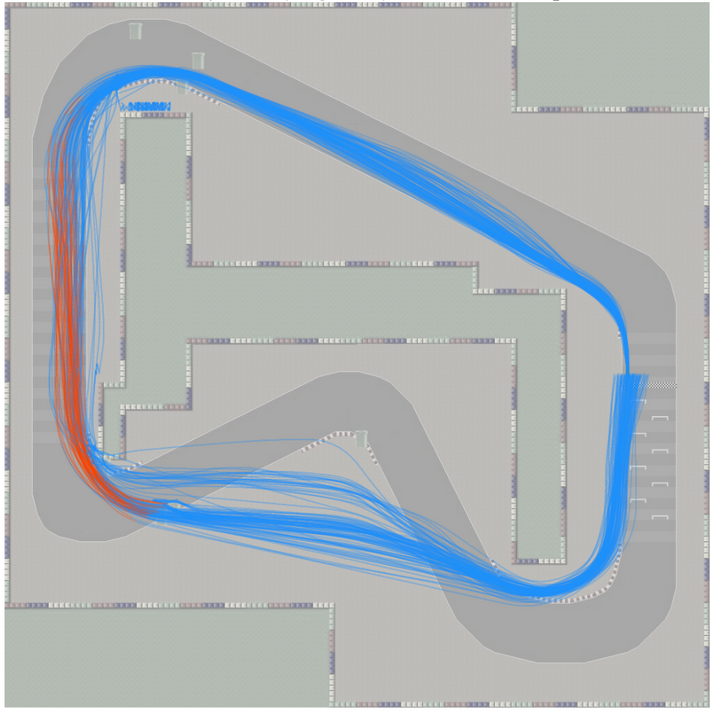}
    \caption{\textbf{Pre-verification model} traces illustrate frequent\textit{$\phi_1)$ Global Speed Limit} and \textit{$\phi_2)$ Stay on Track} rule violations.}
    \label{subfig:mario_before}
  \end{subfigure}
  \hfill
  \begin{subfigure}{0.47\linewidth}
    \centering
    \includegraphics[width=\linewidth]{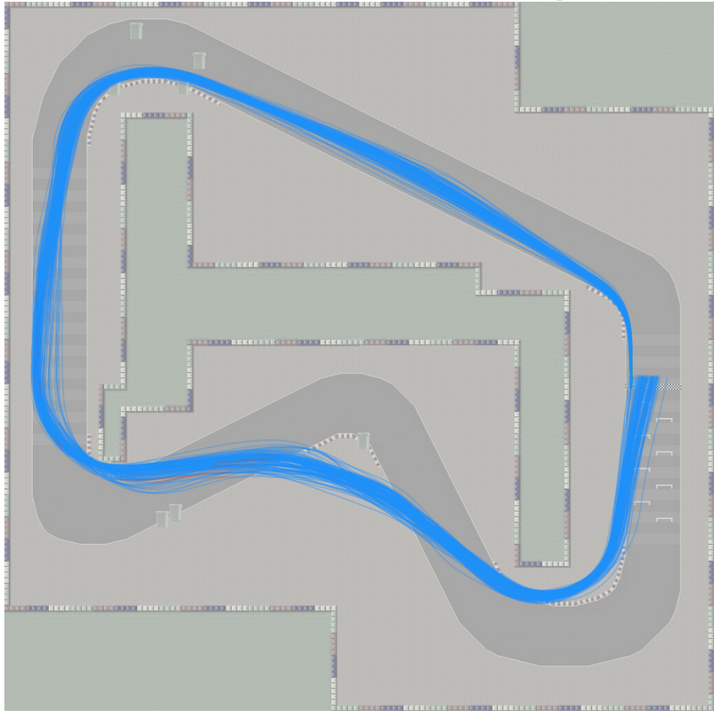}
    \caption{\textbf{Post-verification model} traces show fewer \textit{$\phi_1)$ Global Speed Limit} and \textit{$\phi_2)$ Stay on Track} rule violations.}
    \label{subfig:mario_after}
  \end{subfigure}
  \caption{Traces visualized on the Mario Kart (SNES) platform with red segments marking violations of the $\phi_1)$ \textit{Global Speed Limit} rule. Our approach significantly reduce the number of safety violations in the re-trained model.}
  \label{fig:100traces}
\end{figure}

For evaluation, we rolled out the trained model in an uncluttered, static environment, keeping both goal and environmental conditions fixed. As the policy $\pi_\theta$ is inherently probabilistic, each execution results in a unique set of actions, producing different trace histories even under identical settings. We generated 100 rollout traces (Fig.~\ref{fig:100traces}), capturing variation in the model's behavior across repeated trials. 

\subsection{Mario Kart STL Rules}\label{subsec:mariokart_rules}
We formalize three safety requirements specific to the Mario Kart scenario and present the corresponding STL specifications. These rules correspond to individual formulas $\phi_i$ in the Regulator's specification set $\Phi$.

\vspace{2mm}

\subsubsection{\texorpdfstring{$\phi_1)$}{phi1} Global Speed Limit}
The speed of the car must remain below an upper bound of 90 kph, 

\small{ \begin{equation}
\mathcal{G}_{[0, \infty]}(speed < 90).
\end{equation}}
\normalsize

\subsubsection{\texorpdfstring{$\phi_2)$}{phi2} Stay on Track}
The car must stay on the track and not deviate off course. If the car leaves the track, it must return within 60 timesteps. This represents a track compliance rule with recovery window,

\small{ \begin{equation}
\mathcal{G}_{[0, \infty]} ((surface \equiv \neg track) \rightarrow F_{[0,60]}(surface \equiv track)).
\end{equation}}
\normalsize

\subsubsection{\texorpdfstring{$\phi_3)$}{phi3} Wait to Accelerate}
Once a sharp turn is initiated, the car must not accelerate until the turn rate has substantially stabilized. Here, $\psi$ denotes the heading angle of the car and $\dot{\psi}$ represents its rate of change, the turning velocity. The specification states that when the car has initiated a sharp turn ($\dot{\psi} > 1$), the system must remain in ``not accelerating" ($\neg A$) until it reaches a state where the turn rate has largely leveled out ($\dot{\psi} < 0.04$).

\small{ \begin{equation}
\mathcal{G}_{[0, \infty]} ( (\dot{\psi} > 1) \rightarrow (\neg A \;\; \mathcal{U} \; (\dot{\psi} < 0.04))).
\end{equation}}
\normalsize


\subsection{Model Evaluation and Discussion}

For each STL specification in $\Phi$, we obtain a robustness value for each trace and subsequently compute $TRV(\mathcal{T}_\pi, \phi_i)$, $LRV(\mathcal{T}_\pi, \phi_i)$, $AVRV(\mathcal{T}_\pi,\phi_i)$ as defined in Section~\ref{subsec:regulators}. We use the standard real-valued robustness semantics for STL, where positive values indicate satisfaction and negative values indicate violation. \footnote{We interpret robustness bands qualitatively rather than comparing absolute magnitudes across scenarios. In the Mario Kart scenario, for TRV, we classify values \hlbox{myred}{$<-1$} as overall poor performance, values in \hlbox{myorange}{$[-1,0]$} as marginal performance, and values \hlbox{myyellow}{$>0$} as good performance. For LRV and AVRV, we classify values \hlbox{myred}{$<-10$} as severe violations, values in \hlbox{myorange}{$[-10,-5]$} as moderate violations, and values \hlbox{myyellow}{$>-5$} as mild violations.}

Together with the satisfaction percentage, these metrics, reported in Tables~\ref{table:mario_1}, \ref{table:mario_2}, and~\ref{table:mario_3}, constitute the raw regulator feedback for this scenario. Satisfaction percentage, treated  as a Boolean pass/fail measure, indicates how often specifications were satisfied across the $N$ traces.

\begin{table}[h!]
\centering
\resizebox{0.98\linewidth}{!}{%
\begin{tabular}{|l|c|c|}
\hline
\textbf{Rule 1: Global Speed Limit} & \textbf{pre-verification} & \textbf{post-verification} \\ \hline 
Satisfaction Percentage  &    30\%     &   \textbf{83\%}    \\

TRV      (average performance)  &     \cellcolor{myred}-0.13    &    \cellcolor{myyellow}10.08           \\

LRV      (worst violation)      &      \cellcolor{myred}-12   &    \cellcolor{myorange}-10      \\

AVRV     (average violation)    &     \cellcolor{myred}-11.45      &     \cellcolor{myorange}-5.25  \\
\hline
\end{tabular}
}
\caption{The \textbf{pre-verification model} had frequent and severe violations (low satisfaction, strongly negative AVRV and LRV; more negative is worse). Regulators recommended policy improvement; the \textbf{post-verification model} compliance increased 53\%.}
\label{table:mario_1}
\end{table}

\begin{table}[h!]
\centering
\resizebox{0.98\linewidth}{!}{%
\begin{tabular}{|l|c|c|}
\hline
\textbf{Rule 2: Stay on Track} & \textbf{pre-verification} & \textbf{post-verification} \\ \hline 
Satisfaction Percentage  &  8\%      &   \textbf{99\% }     \\

TRV    (average performance)     &  \cellcolor{myred}-17.4  &   \cellcolor{myyellow}0.8      \\

LRV    (worst violation)         &  \cellcolor{myred}-19    &  \cellcolor{myred}-19       \\

AVRV     (average violation)    &     \cellcolor{myred}-19      &     \cellcolor{myred}-19  \\
\hline
\end{tabular}
}
\caption{The \textbf{pre-verification model} was almost always in severe violation of this specification (low TRV, LRV, and AVRV). Regulators recommended policy improvement; the \textbf{post-verification model} compliance increased by 91\%.}
\label{table:mario_2}
\end{table}

\begin{table}[h!]
\centering
\resizebox{0.98\linewidth}{!}{%
\begin{tabular}{|l|c|c|}
\hline
\textbf{Rule 3: Wait to Accelerate} & \textbf{pre-verification} & \textbf{post-verification} \\ \hline \
Satisfaction Percentage        &    87\%    &   \textbf{95\%}     \\

TRV     (average performance)  &   \cellcolor{myorange}-0.06    &   \cellcolor{myyellow}0.55     \\

LRV     (worst violation)      &   \cellcolor{myyellow}-2.14    &   \cellcolor{myyellow}-2.14        \\

AVRV   (average violation)     &   \cellcolor{myyellow}-1.56 &     \cellcolor{myyellow}-2.13 \\
\hline
\end{tabular}
}
\caption{The \textbf{pre-verification model} was largely satisfied with mild/moderate violations. Regulators recommended prioritizing other rules.}
\label{table:mario_3}
\end{table}

The results illustrate that $\phi_2)$ \textit{Stay on Track} presented the greatest challenge with only 8\% satisfaction, the lowest TRV, and the most severe violations. $\phi_1)$ \textit{Global Speed Limit} rule also showed low compliance (30\% satisfaction). By contrast, $\phi_3)$ \textit{Wait to Accelerate} rule was largely satisfied in the pre-verification model, with relatively mild violations. Regulators aggregated these evaluation outcomes into a safety score $S(\pi)$, incorporating their domain-informed specification weighting, and communicated this assessment to designers to guide policy retraining.

In response, Designers modified the reward structure to train a more compliant agent. A larger penalty was applied when the car left the track for extended periods of time, and a speeding penalty was reinforced. The resulting post-verification reward structure is

\begin{equation} \label{eq:mario_reward_after}
  R_{post} = R_{cp} + R_{motion} + \uparrow R_{road} + \uparrow R_{speed},
\end{equation}

reflecting a Designer update that shifts the balance between task reward and the safety score $S(\pi)$.


\section{Mobile Robot Navigation}  \label{sec:turtlebots}

\begin{figure}[htbp]
  \centering
  \begin{subfigure}{0.46\linewidth}
    \centering
    \includegraphics[width=\linewidth]{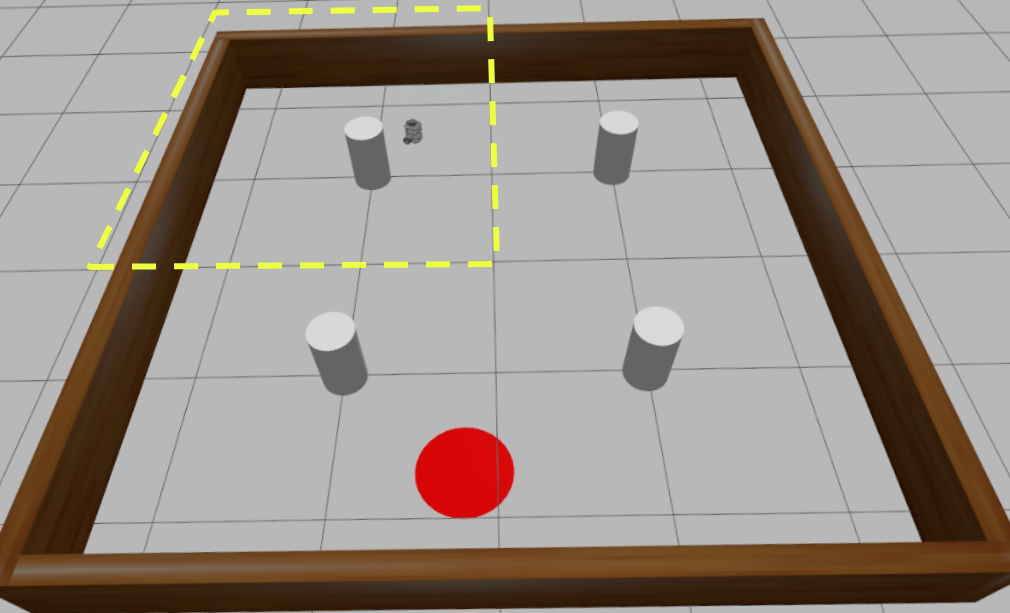}
    \caption{The gazebo simulation environment used to train TurtleBot3 ($5 \times5$m). The dashed yellow line indicates equivalent area in (\subref{subfig:real})}
    \label{subfig:gazebo}
  \end{subfigure}
  \hfill
  \begin{subfigure}{0.40\linewidth}
    \centering
    \includegraphics[width=\linewidth]{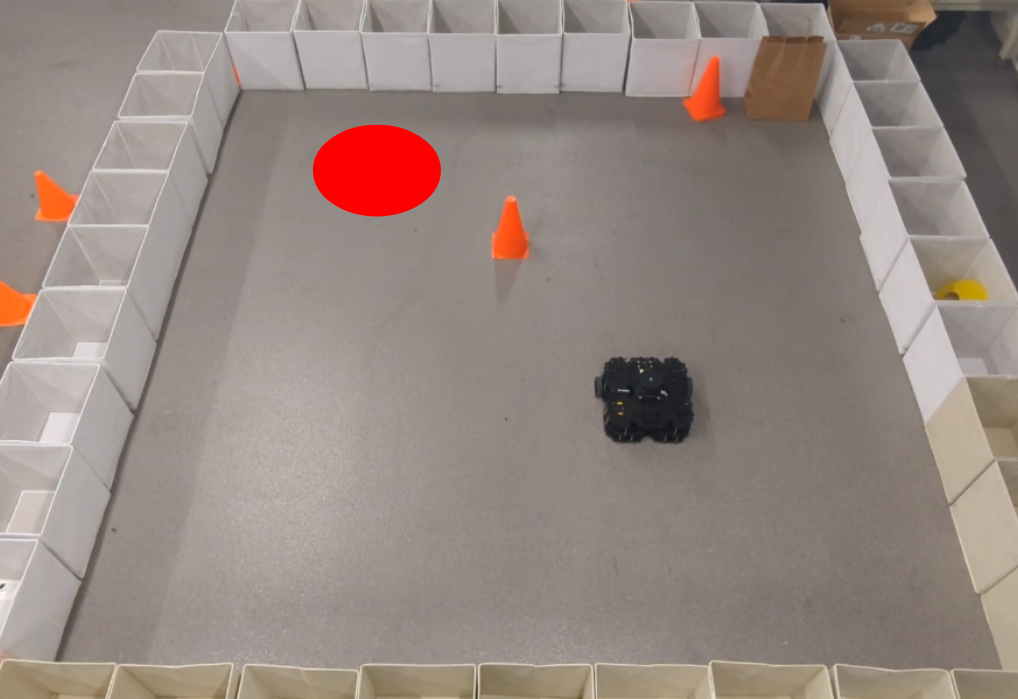}
    \caption{Real world demonstration area ($2.5 \times 2.5$m) for the turtlebot3, analogous to the yellow highlighted area in (\subref{subfig:gazebo}).}
    \label{subfig:real}
  \end{subfigure}
  \caption{TurtleBot3 scenarios in simulation (\subref{subfig:gazebo}) and real-world (\subref{subfig:real}). The red circles indicate approximate goal locations.}
  \label{fig:gazebo}
\end{figure}

The Mobile Robot Navigation scenario uses the \textbf{Turtlebot3} platform due to its low cost, strong ROS2 integration, and easily available documentation. We adapt an existing scenario from the Turtlebot3 manual~\cite{turtlebot3} to train our black-box machine-learning model. In this setup, the TurtleBot must reach a randomly selected target point on the map while avoiding obstacles (Fig. \ref{fig:gazebo}). The robot has a simplified action space, selecting from five discrete angular velocities while its linear velocity is held constant. At each step, the TurtleBot's observation includes 24 lidar points covering $180$ degree field of view along with the heading and distance to the goal. 

The reward structure combines a yaw reward that encourages goal alignment ($R_{raw}$) and an obstacle reward that penalizes proximity to obstacles $R_{obstacle}$). The robot also receives a large positive or negative terminal reward ($R_{terminal}$) at the end of each episode corresponding to success or failure. The pre-verification reward can be summarized as below (Eq. \ref{eq:turtlebot_reward}),

\begin{equation} \label{eq:turtlebot_reward}
R_{pre} = R_{yaw} + R_{obstacle} + R_{terminal}.
\end{equation}

The original,\textbf{ pre-verification model} trained in the TurtleBot3 environment for 500 episodes, depicted in Figure \ref{fig:turtlebot_before}. This pre-verification model represents a black-box controller that is under-trained and exhibits unsafe behavior. 

\subsection{TurtleBot STL Rules}\label{subsec:turtlebot_rules}

We derive the following human-readable rules and their corresponding STL specifications.

\vspace{2mm}
\subsubsection{\texorpdfstring{$\phi_1)$}{phi1} No Sharp Turns}
The TurtleBot must avoid long, sharp turns to minimize wheel wear and tear. Here, $\psi$ denotes the heading angle of the robot, and $\dot{\psi}$ represents its rate of change, the turning velocity. The specification states that if the TurtleBot turns sharply ($\dot{\psi}>0.2$), then within 50 timesteps, it must reduce its turning rate to a gentle level ($\dot{\psi} \leq 0.2$) and maintain that for at least 5 timesteps.

\vspace{2mm}
\small{ \begin{equation}
G_{[0, \infty]} ((|\dot{\psi}| > 0.2) \rightarrow F_{[0,50]} (G_{[0,5]} (|\dot{\psi}| \leq 0.2))).
\end{equation}}
\normalsize

\subsubsection{\texorpdfstring{$\phi_2)$}{phi2} Timed Completion}
The TurtleBot must reach the goal successfully within 800 timesteps,

\small{ \begin{equation}
F_{[0, 800]} (goal\_reached = True).
\end{equation}}
\normalsize

\subsubsection{\texorpdfstring{$\phi_3)$}{phi3} Don't Linger Near an Obstacle}
If the TurtleBot moves to within 0.5 meters of an obstacle, it must move away within 50 timesteps,

\small{ \begin{equation}
G_{[0, \infty]} ((dist_{obst} < 0.5 \; m) \rightarrow F_{[0,50]}(dist_{obst} \geq 0.5 \; m)).
\end{equation}}
\normalsize

\subsection{Model Evaluation and Discussion}
Robustness bands are defined separately for the mobile robot navigation scenario, where robustness has no canonical scale across different environments and formulas. \footnote{For TRV, we classify values \hlbox{myred}{$<-0.2$} as overall poor performance, values in \hlbox{myorange}{$[-0.2,-0.01])$} as marginal performance, and values \hlbox{myyellow}{$>-0.01$} as good performance. For LRV and AVRV, we classify values \hlbox{myred}{$<-0.5$} as severe violations, values in \hlbox{myorange}{$[-0.5,-0.1]$} as moderate violations, and values \hlbox{myyellow}{$>-0.1$} as mild violations.}

\begin{table}[h!]
\centering
\resizebox{0.98\linewidth}{!}{%
\begin{tabular}{|l|c|c|}
\hline
\textbf{Rule 1: No Sharp Turns} & \textbf{Pre-Verification}& \textbf{Post-Verification}\\ 
\hline 
Satisfaction Percentage  &      9\%   &      \textbf{36\%}      \\

TRV (average performance) &      \cellcolor{myorange}{-0.01}  &     \cellcolor{myyellow}{-0.005}    \\

LRV (worst violation)     &      \cellcolor{myyellow}{-0.03}  &     \cellcolor{myyellow}{-0.03}  \\ 

AVRV (average violation) & \cellcolor{myyellow}{-0.01} & \cellcolor{myyellow}{-0.01} \\
\hline
\end{tabular}
}
\caption{The \textbf{pre-verification model} had low satisfaction, with frequent but very mild violations (negative but small TRV, LRV, and AVRV). Regulators recommended policy improvement; the \textbf{post-verification model} compliance increased by 53\%.}
\label{table:turtle1}
\end{table}

\begin{table}[h!]
\centering
\resizebox{0.98\linewidth}{!}{%
\begin{tabular}{|l|c|c|}
\hline
\textbf{Rule 2: Timed Completion} & \textbf{Pre-Verification}& \textbf{Post-Verification}\\ 
\hline 
Satisfaction Percentage    &  18\%        &  \textbf{54\%}       \\

TRV (average performance) &  \cellcolor{myred}{-0.40}     &  \cellcolor{myorange}{-0.14}    \\

LRV (worst violation)     &   \cellcolor{myred}{-1.66}    &  \cellcolor{myred}{-0.88} \\

AVRV (average violation ) & \cellcolor{myorange}{-0.48} & \cellcolor{myorange}{-0.31} \\
\hline
\end{tabular}
}
\caption{The \textbf{pre-verification model} had low satisfaction and clear negative TRV and AVRV, suggesting that the model systematically failed to finish on time. Regulators recommended policy improvement; the \textbf{post-verification model} compliance increased by 36\%.}
\label{table:turtle2}
\end{table}

\begin{table}[!ht]
\centering
\resizebox{0.98\linewidth}{!}{%
\begin{tabular}{|l|c|c|}
\hline
\textbf{Rule 3: Don't Linger} & \textbf{Pre-Verification}& \textbf{Post-Verification}\\ 
\hline 
Satisfaction Percentage    &       45\%   &     \textbf{67\%}     \\

TRV (average performance) &    \cellcolor{myyellow}{0.02}    &    \cellcolor{myyellow}{0.1}    \\

LRV (worst violation)     &   \cellcolor{myorange}{-0.23}    &    \cellcolor{myorange}{-0.22}    \\

AVRV (average violation ) & \cellcolor{myorange}{-0.14} & \cellcolor{myorange}{-0.13} \\
\hline
\end{tabular}
}
\caption{The \textbf{pre-verification model} exhibited mixed behavior with balanced pass/fails and non-catastrophic violations. Regulators de-prioritized this rule in their recommendations.}
\label{table:turtle3}
\end{table}

\begin{figure}
    \centering
  \begin{subfigure}{0.45\linewidth}
    \centering
    \includegraphics[width=\linewidth]{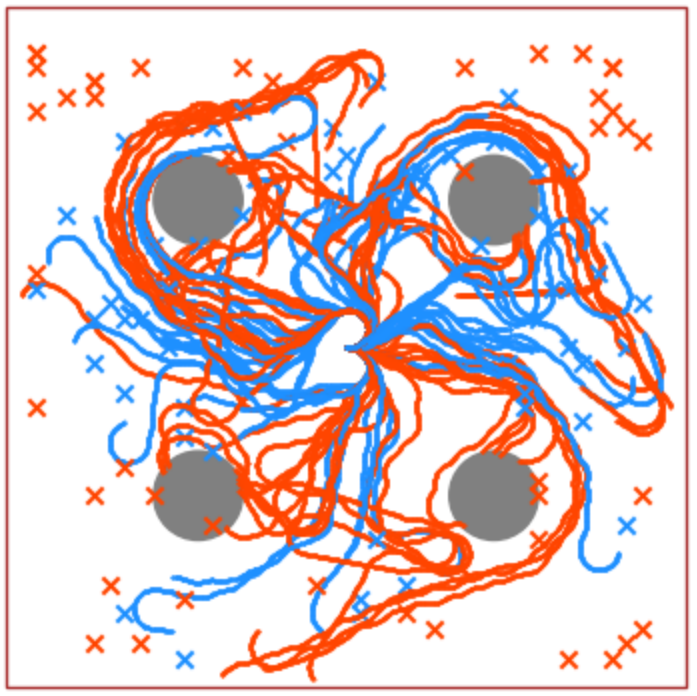}
    \caption{\textbf{Pre-verification} model successfully reaches the goal in \textbf{53} / 100 traces.}
    \label{fig:turtlebot_before}
  \end{subfigure}
  \hfill
  \begin{subfigure}{0.45\linewidth}
    \centering
    \includegraphics[width=\linewidth]{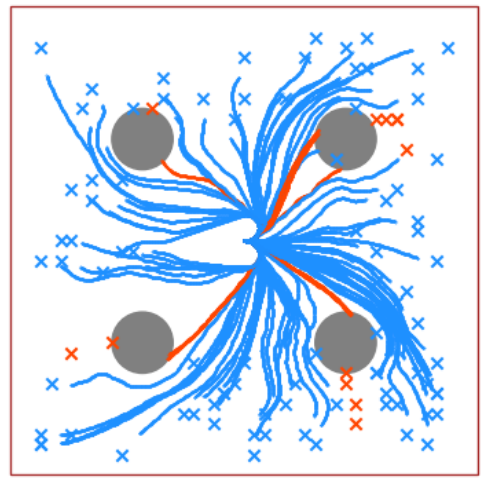}
    \caption{\textbf{Post-verification} model successfully reaches the goal in \textbf{87} / 100 traces.}
    \label{fig:turtlebot_after}
  \end{subfigure}
    \caption{TurtleBot3 traces visualized in the environment with static obstacles (gray circles) and randomly generated goals (crosses). The robot is initialized in the center of the map. Trajectories are shown as lines illustrating timeouts or collisions (red) and successes (blue).}
\end{figure}

Applying the recommendation criteria from Section \ref{subsec:recommendations}, the \textbf{pre-verification model} triggered policy-improvement recommendations for all three rules. Observing the pre-verification metrics in Tables \ref{table:turtle1}, \ref{table:turtle2}, and \ref{table:turtle3}, Regulators noted particularly poor adherence to $\phi_1)$ \textit{No Sharp Turns} rule (9\% satisfaction) and $\phi_3)$ \textit{Don't Linger Near an Obstacle} (18\% satisfaction). The raw scores ($TRV(\mathcal{T}_\pi,\phi_i)$, $LRV(\mathcal{T}_\pi,\phi_i)$, and $AVRV(\mathcal{T}_\pi,\phi_i)$) were combined with domain-expertise in the scoring function and conveyed to Designers as an overall safety score. As one way to act on this feedback, Designers introduced a small penalty on changes in angular acceleration ($R_\alpha$) to encourage smoother turning, and increased the penalty associated with close proximity to obstacles ($R_{obstacle} \uparrow$), encouraging the policy to avoid lingering near them. The resulting post-verification reward structure is (Eq. \ref{eq:turtlebot_reward_after}),

\begin{equation} \label{eq:turtlebot_reward_after}
R_{post} = R_{yaw} + \uparrow \!\! R_{obstacle}  + R_{terminal} + R_{\alpha}.
\end{equation}  

After a few hundred additional training episodes, rollout traces from the \textbf{post-verification model} (Fig. \ref{fig:turtlebot_after}) depict more direct paths to goals, and the metrics in Tables \ref{table:turtle1}, \ref{table:turtle2}, and \ref{table:turtle3} indicate improved adherence to all three specifications.

\subsection{Real World Experiments} \label{sec:real_world_exp}
\begin{figure}[h!]
    \centering
  \begin{subfigure}{0.48\linewidth}
    \centering
    \includegraphics[width=\linewidth]{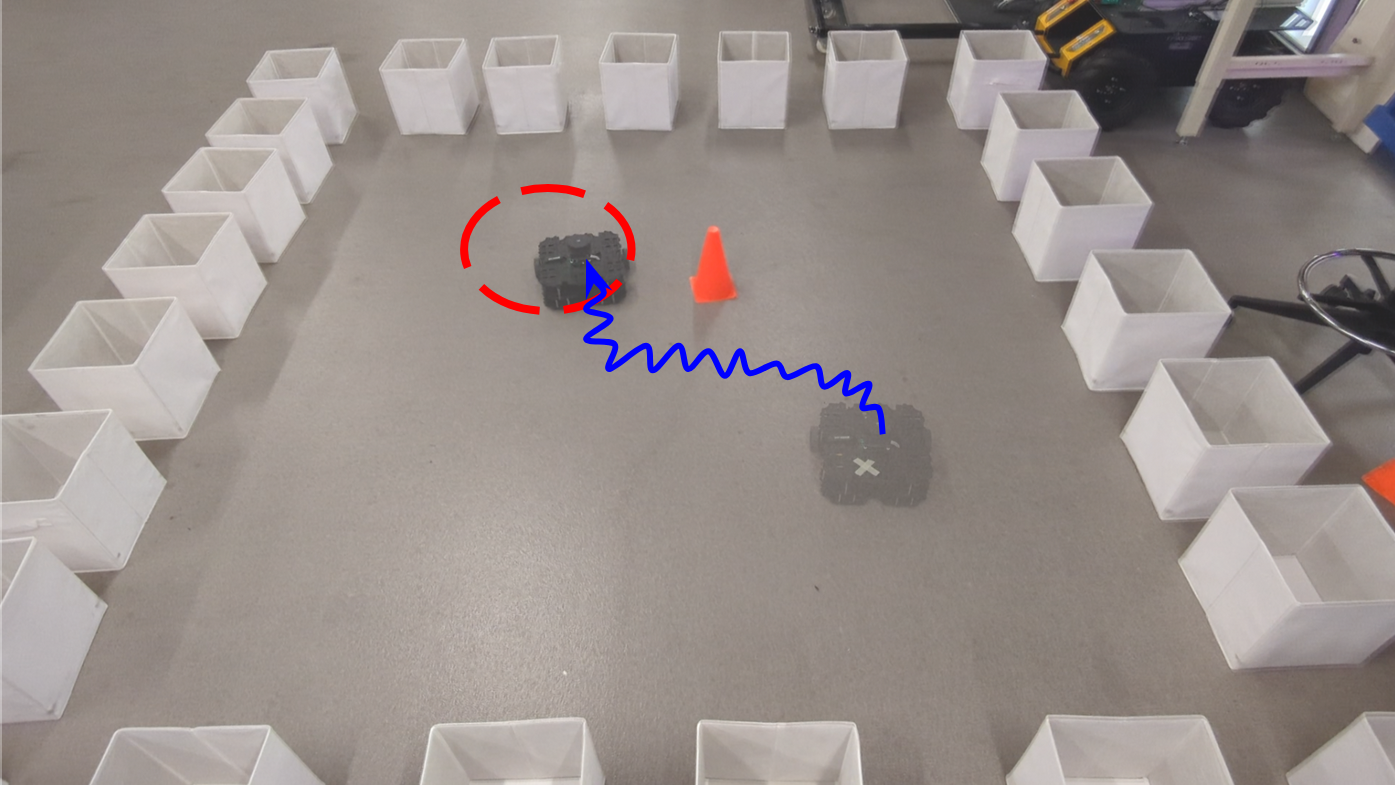}
    \caption{Original model pre-verification}
    \label{subfig:real_robot_before}
  \end{subfigure}
  \hfill
  \begin{subfigure}{0.48\linewidth}
    \centering
    \includegraphics[width=\linewidth]{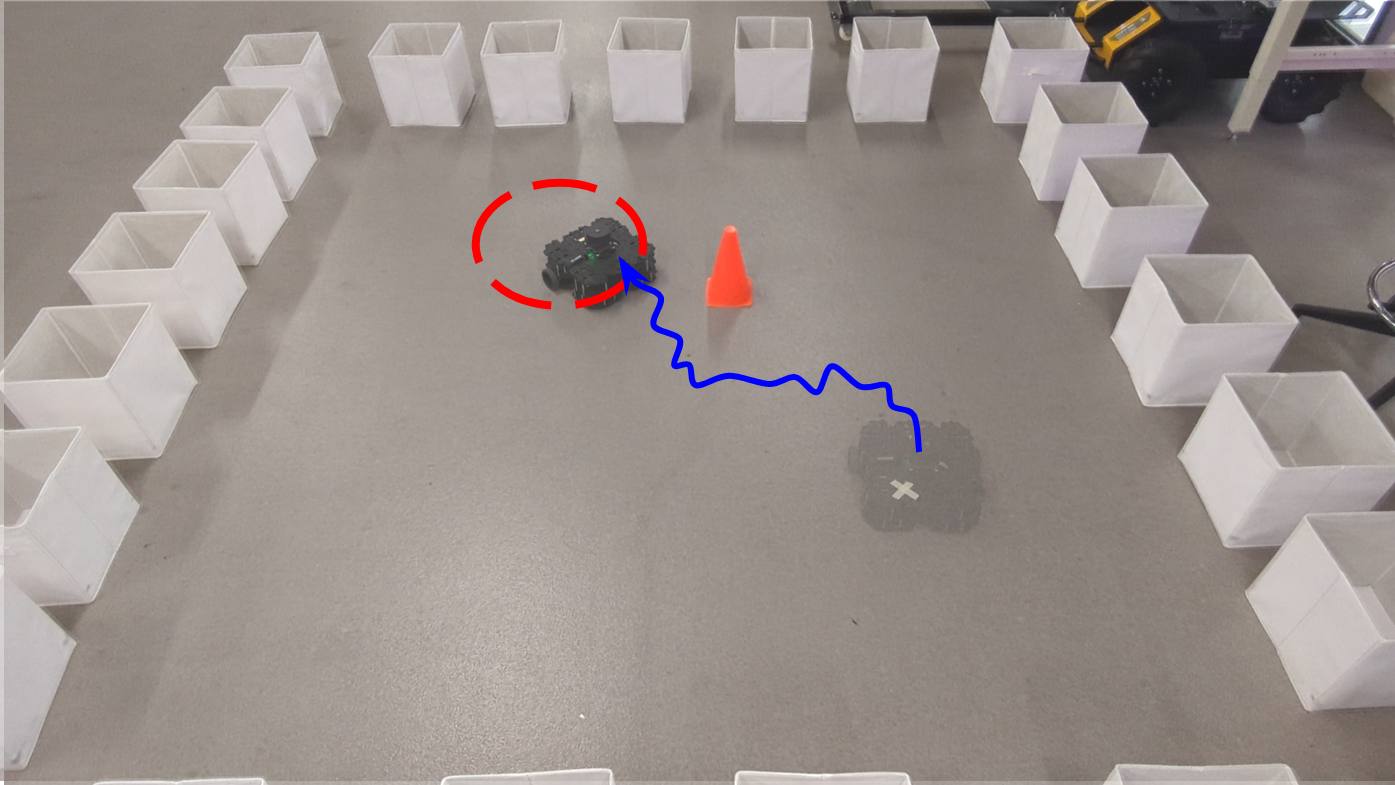}
    \caption{Fine-tuned model post-verification}
    \label{subfig:real_robot_after}
  \end{subfigure}
    \caption{Real-world demonstrations of TurtleBot3 pre and post safety verification. The red circle denotes the goal and the blue line is the trajectory taken by the robots (manually annotated). The post verification model showed a smoother path, even if the path demonstrated is not as smooth as the simulation paths.}
    \label{fig:real_robot_before_after}
\end{figure}

Both the TurtleBot3 \textbf{pre-verification model} and \textbf{post-verification model} were deployed in a real-world proxy of our simulation environment. To accommodate the area restriction of an indoor testing environment (as seen in Fig. \ref{subfig:gazebo}, \ref{subfig:real}), the real world demonstration was scaled down to 25\% of the original scenario, and the robot's speed was halved to safely navigate a smaller map. As observed in Figure \ref{fig:real_robot_before_after}, while both pre- and post-verification policies reach the goal, the post-verification policy demonstrated a comparatively smoother path. We observe a sim-to-real gap: the real robot makes more frequent turns in real-world (Fig. \ref{subfig:real_robot_after}) compared to the trajectories seen in simulation (Fig. \ref{fig:turtlebot_after}). However, we still note the improved post-verification model, demonstrating our verification workflow improves robot behavior in real-world demonstrations.

\section{Conclusions, Limitations, and Future Work}
\ours{} presents a step toward formal verification of black-box robot policies using temporal safety specifications. By evaluating traces alone, our approach respects the constraints of real-world certification processes, where autonomy stacks are often opaque and only input-output behavior is observable. During evaluation, we formalize time-dependent safety requirements as STL specifications, enabling domain experts to formalize temporal safety requirements that are rarely accessible in black-box verification settings.  

We use robustness-based metrics (TRV, LRV, and AVRV) to derive quantitative safety scores that summarize average performance, worst-case behavior, and violation severity over time. Combined with simple decision rules, these metrics allow the regulator to provide targeted, specification-level feedback and concrete retraining recommendations, rather than relying solely on aggregate failure rates or surrogate safety indicators. Our experiments across two domains and six unique specifications demonstrate that this process can substantially increase specification satisfaction rates and improve robustness metrics for deployed learning-based policies.

Limitations of our approach include a lengthy iterative process between the regulator and designer and ambiguous translation of specifications by human regulators. Future work aims to incorporate LLM-based translation tools to standardize STL specification translation from natural language rules. Although our approach does not yet address coverage or rare-event analysis, it is designed to complement rather than replace policy synthesis, and its STL specifications should be accessible to designers to promote shared interpretability and collaborative refinement.

\bibliographystyle{IEEEtran}
\bibliography{ref_arxiv}

\end{document}